\documentclass{article}
\usepackage{spconf,amsmath,graphicx,booktabs}
\usepackage{multirow}
\usepackage{subfigure} 
\usepackage{epstopdf}
\usepackage{pgfplots}
\pgfplotsset{compat=1.9}
\usepackage{hyperref}
\usepackage{url}
\graphicspath{{figs/}}
\usepackage{setspace}

\title{A Token-level Contrastive Framework for Sign Language Translation}
\name{Biao Fu$^{1,2,}$\sthanks{\,\, Equal contribution.} 
Peigen Ye$^{1,2,}$\footnotemark[1] 
Liang Zhang$^{1,2}$ 
Pei Yu$^{1,2}$ 
Cong Hu$^{1,2}$ 
Xiaodong Shi$^{1,2}$ 
Yidong Chen$^{1,2,}$\sthanks{\,\, Corresponding author.}
} 
\address{$^{1}$Department of Artificial Intelligence, School of Informatics, Xiamen University\\
$^{2}$Key Laboratory of Digital Protection and Intelligent Processing of Intangible \\ Cultural Heritage of Fujian and Taiwan, Ministry of Culture and Tourism}

\begin{document}
%
\maketitle
\begin{abstract}
Sign Language Translation (SLT) is a promising technology to bridge the communication gap between the deaf and the hearing people. Recently, researchers have adopted Neural Machine Translation (NMT) methods, which usually require large-scale corpus for training, to achieve SLT.
However, the publicly available SLT corpus is very limited, which causes the collapse of the token representations and the inaccuracy of the generated tokens.
To alleviate this issue, we propose ConSLT, a novel token-level \textbf{Con}trastive learning framework for  \textbf{S}ign \textbf{L}anguage \textbf{T}ranslation
, which learns effective token representations by incorporating token-level contrastive learning into the SLT decoding process.
Concretely, ConSLT treats each token and its counterpart generated by different dropout masks as positive pairs during decoding, and then randomly samples $K$ tokens in the vocabulary that are not in the current sentence to construct negative examples.
We conduct comprehensive experiments on two benchmarks (PHOENIX14T and CSL-Daily) for both end-to-end and cascaded settings.
The experimental results demonstrate that ConSLT can achieve better translation quality than the strong baselines \footnote{\url{https://github.com/biaofuxmu/ConSLT}}. 
\end{abstract}

\begin{keywords}
Sign language translation, low-resource, contrastive learning
\end{keywords}

\section{Introduction}
\label{sec:intro}
Sign Language Translation (SLT), which takes a sign video as the input and generates a spoken language sentence (or text), can play an important role in facilitating the communication between the deaf and the hearing people.
Thus, SLT has attracted great attention from the research community \cite{camgoz2018neural,camgoz2020multi,zhou2021improving,zhou2021spatial,yin-etal-2021-including,chen2022simple}. %
Existing works \cite{camgoz2018neural,camgoz2020multi,camgoz2020sign,yin-read-2020-better,xie2021pisltrc} treated SLT as a NMT problem and adopted NMT methods, which usually require large-scale corpus for training, to achieve SLT. 
However, the most popular SLT dataset PHOENIX14T \cite{camgoz2018neural} only contains less than 9K parallel data, whose size is several orders of magnitude smaller than the NMT datasets, as shown in Table \ref{tab:statistics}. 
Unfortunately, it is not easy for SLT systems to leverage large-scale annotation data for training deep models like NMT, because
the collection and annotation of the sign language datasets are extremely difficult and expensive. 
Thus, the SLT task is essentially a low-resource problem \cite{yin-etal-2021-including}.

\begin{figure}[!tbp]
\centering
\subfigure[STMC-Transformer]{
\label{fig:visual_stmc}
\includegraphics[scale=0.28]{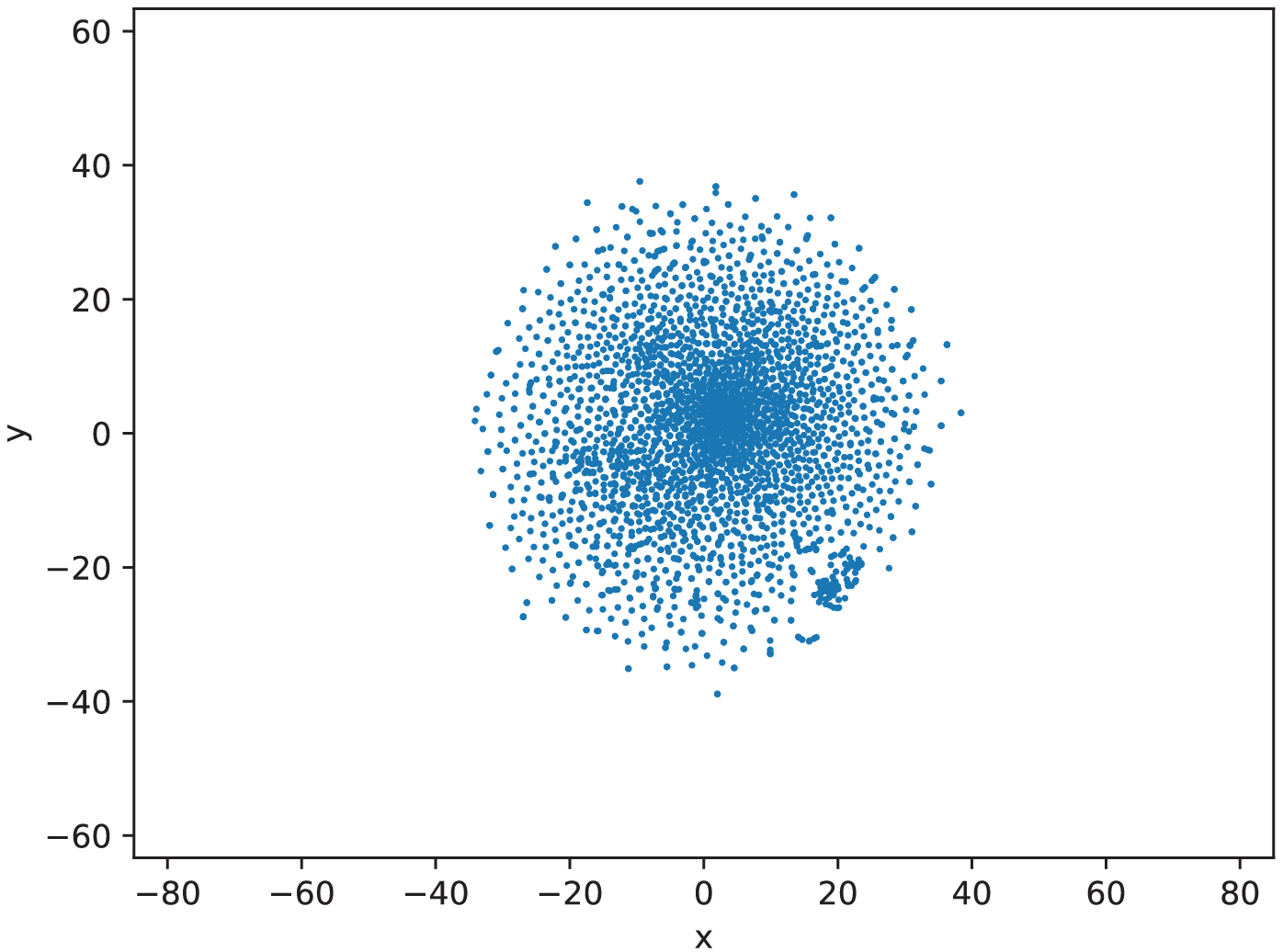}
}
~
\subfigure[ConSLT]{
\label{fig:visual_conslt}
\includegraphics[scale=0.28]{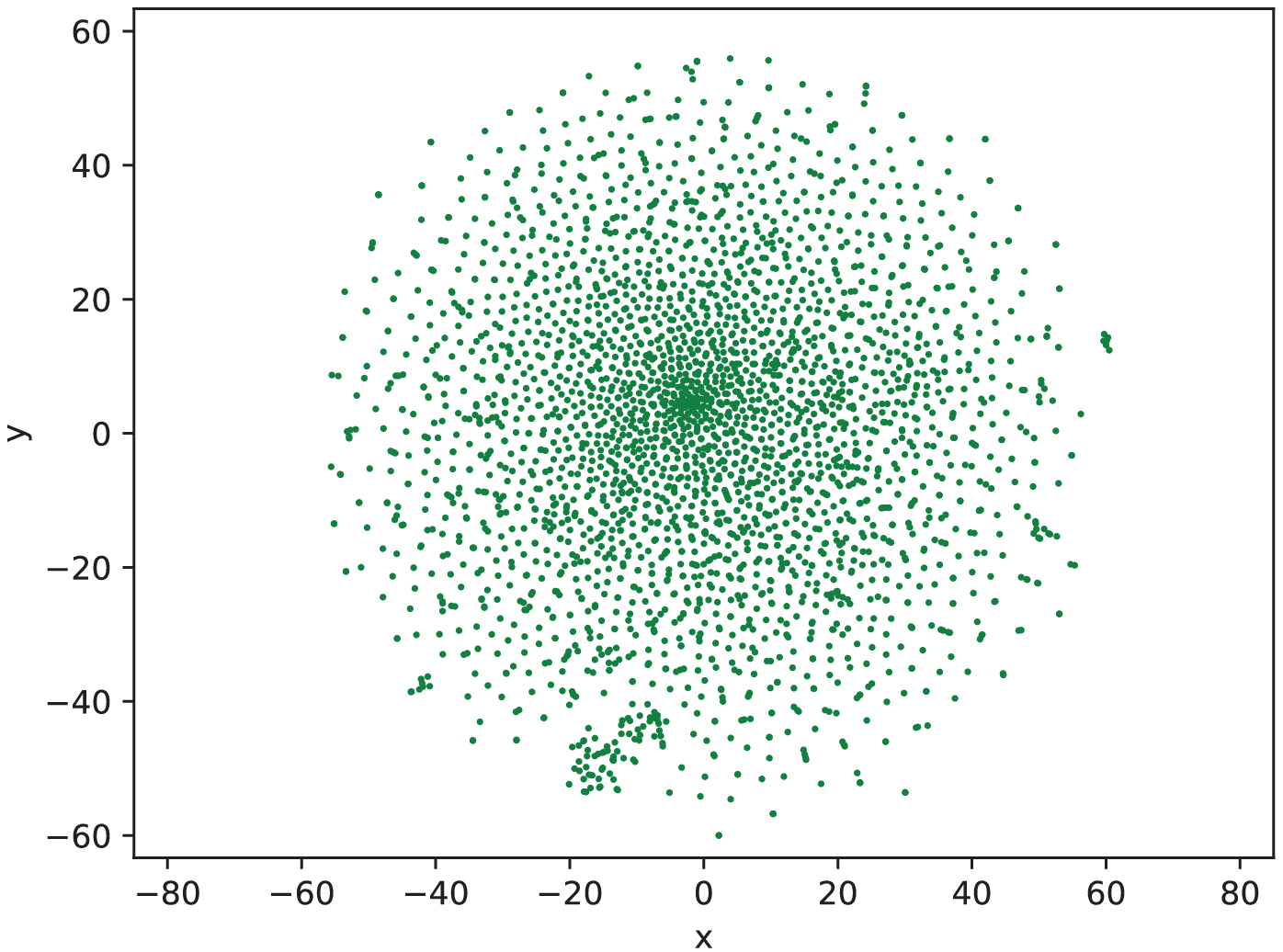}
}
\caption{Visualization of token embeddings learned by (a) STMC-Transformer and (b) ConSLT. The visualization is generated by employing t-SNE \cite{vandermaaten2008tsne}, where each point represents a token from the vocabulary.}
\label{fig:visualization}
\end{figure}

\begin{table}[!tbp]
\centering
\begin{tabular}{l|c|cc}
\toprule
Corpus & PHOENIX14T & \multicolumn{2}{c}{WMT18 En-De} \\ \cmidrule(lr){1-4}
Sentences & 8,257  & \multicolumn{2}{c}{1,920,209}                \\ \cmidrule(lr){1-4}
Words     & 72,783 & \multicolumn{1}{c|}{53,008,851} & 50,486,398 \\ \bottomrule
\end{tabular}
\caption{Statistics of SLT and NMT dataset. The size of SLT dataset PHOENIX14T is several orders of magnitude smaller than NMT dataset WMT18 En-De \cite{bojar-etal-2018-findings}.}
\label{tab:statistics}
\end{table}
Recent works \cite{li2020prototypical,pan-etal-2021-contrastive,wei2021learning} have demonstrated that contrastive learning has tremendous potential in low-resource scenarios, including low-resource NMT 
 \cite{pan-etal-2021-contrastive,wei2021learning}. 
However, these efforts normally aim to obtain better sentence representations, while the translation quality of NMT is more related to the accuracy of the tokens since NMT systems predict each token based on the previously generated tokens. Moreover, it could be observed
that data scarcity of NMT (SLT in our case) really leads to poor token representations.
As shown in Figure \ref{fig:visual_stmc}, the majority of token embeddings are squeezed into a narrow space (
representation collapse phenomenon \cite{gao2018representation}
), resulting in indistinguishable token representations.

To alleviate this issue, we propose a novel token-level contrastive learning framework for SLT, called ConSLT.
Specifically, for each token at each decoding step, ConSLT takes it and its augmented versions generated by dropout noises as positive pairs.
Then, instead of using all other in-batch sentences or tokens as the negatives \cite{gao2021simcse,yan-etal-2021-consert,zhang2022frequency}, ConSLT randomly samples $K$ tokens in vocabulary that are not in the current sentence as its negative examples to mine diverse negative instances.
As a comparison shown in Figure \ref{fig:visual_conslt},
ConSLT produces a more uniform representation space and successfully solve the collapse issue caused by SLT data scarcity to generate more accurate translations.

In addition, our method is model-agnostic and can be adapted to different model configurations.
Thus, we evaluate ConSLT on the two challenging datasets, i.e., PHOENIX14T and CSL-Daily, for end-to-end and cascaded settings.
We also explore and analyze several negative sampling strategies for ConSLT.
Extensive experiments demonstrate that our method can effectively improve the translation quality of SLT. 

\section{Related Works}
\textbf{Sign Language Translation}.  \cite{camgoz2018neural} first formulated SLT as a NMT problem and adopted NMT framework to solve the SLT problem. 
Then, the subsequent works introduced new technologies or new information to further improve the translation quality of the systems,
including multi-task learning \cite{camgoz2020sign},
data augmentation \cite{zhou2021improving}, hierarchical neural network \cite{li2020tspnet,kan2022sign},
multi-cue features (i.e., hand shapes, facial expressions, and body postures) \cite{camgoz2020multi,zhou2021spatial}, 
position information \cite{xie2021pisltrc},
and signer-independent settings \cite{jin2021contrastive}.
Although the above works have achieved varying degree of progress, most of them only focus on the improvement of the visual feature module.
However, NLP techniques are crucial to improving the translation quality of SLT \cite{yin-etal-2021-including}.
In this paper, we expect to learn effective token representations for SLT by introducing contrastive loss in the translation stage to tackle the low-resource problem of SLT.

\textbf{Contrastive Learning}, which aims to learn effective representations by pulling the positive pairs close and pushing the negative pairs away,  has been widely used in language modeling \cite{gao2021simcse,yan-etal-2021-consert}, text summarization\cite{liu-liu-2021-simcls}, and NMT \cite{pan-etal-2021-contrastive,wei2021learning,zhang2022frequency}.
\cite{gao2021simcse} apply the dropout twice acting as data augmentation to obtain two different sentence representations as positive pairs for each sentence. 
\cite{wei2021learning} propose word-level contrastive learning to learn universal sentence representations across languages by distinguishing words semantically related to sentences. 
These works aim to learn better sentence representations by contrastive learning. 
The closest work to ours is \cite{zhang2022frequency}, which focuses  on low-frequency word prediction in NMT by leveraging token-level contrastive learning. 
However, ConSLT learns effective token representations to alleviate the low-resource problem for SLT.
Moreover, ConSLT can mine diverse negative examples by randomly selecting $K$ tokens in the vocabulary that are not in the current sentence, thereby achieving better translation quality (Table \ref{tab:sampling}).

\section{Approach}
\label{sec:method}
\subsection{Problem Formulation}
Existing SLT corpora are composed of video-text pairs $\left(\boldsymbol{x},\boldsymbol{y}\right)$, where $\boldsymbol{x}=\left(x_{1}, \ldots, x_{T_x}\right)$ is a sign video with $T_x$ frames and 
$\boldsymbol{y}=\left(y_{1}, \ldots, y_{T_y}\right)$ denotes the corresponding spoken sentence.
SLT models aim to translate a sign video $\boldsymbol{x}$ into spoken text $\boldsymbol{y}$.
Existing SLT systems can be categorized into two setting.
(\romannumeral1) \textit{Sign2Text}: an end-to-end setting, which translates sign videos directly into texts.
(\romannumeral2) \textit{Sign2Gloss2Text}: a cascaded setting (or two-stage pipeline), which first predicts the gloss sequences $\boldsymbol{z}=\left(z_{1}, \ldots, z_{T_z}\right)$ with a sign language recognition model and then translates the predicted glosses into texts with a translation model.
ConSLT can be adapted to these two settings due to its model-agnostic nature.

\begin{figure}[!tbp]
\centering
\includegraphics[scale=0.56]{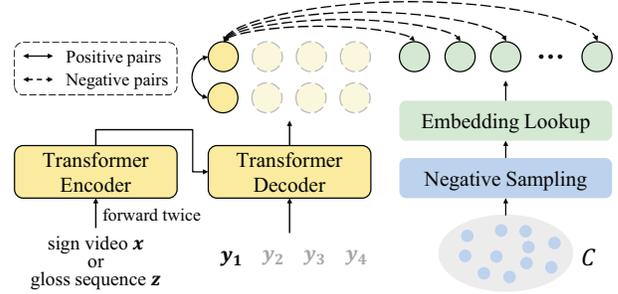}
\caption{Illustration of the ConSLT. For each token, we construct its positive examples by different dropout noises and randomly sample $K$ tokens from a candidate token set $\mathcal{C}$ as negative examples, where $\mathcal{C} \subset \mathcal{V}\backslash \mathcal{S}$ denotes the tokens that are in vocabulary $\mathcal{V}$ but are not in the current sentence $\mathcal{S}$.} 
\label{fig:arch}
\end{figure}

\subsection{Token-level Contrastive Learning}
\label{sec:tcl}
The overview of ConSLT is illustrated in Figure \ref{fig:arch}. 
Following \cite{camgoz2020sign,yin-read-2020-better}, ConSLT adapts vanilla Transformer \cite{vaswani2017attention} structure.
We use a pre-trained video feature extractor \cite{camgoz2020sign}, which is frozen during training, to embed sign video frames. 
Details for the construction of positive pairs, negative pairs, and the distance metrics are described as follows. \\
\textbf{Positive pairs}. 
Inspired by SimCSE\cite{gao2021simcse}, we feed a sign video $\boldsymbol{x}$ (the input of Sign2Text) or a sign gloss sequence $\boldsymbol{z}$ (the input of Sign2Gloss2Text) to the model twice by employing different dropout noises. 
Since the dropout mechanism randomly drops part of units in the Transformer, we can obtain two different hidden representations as the positive pairs for each token $y_t$, denoted as ($h_t, h_{t}^{+}$).
\\
\textbf{Negative pairs}. It is a natural choice to use all other in-batch tokens as the negatives \cite{zhang2022frequency}. 
However, this approach has two major flaws. 
First, the diversity of the tokens in the mini-batch is insufficient.
Second, different tokens from the same sentence cannot be pushed apart due to the contextual dependency.
To this end, we propose a novel negative sampling strategy.
Given a sentence $\boldsymbol{y}$, we denote the set of all tokens in $\boldsymbol{y}$ as $\mathcal{S}$. 
We first randomly sample $K$ tokens from a candidate token set $\mathcal{C} \subset \mathcal{V} \backslash \mathcal{S}$ (all tokens that are in vocabulary $\mathcal{V}$ but are not in the current sentence $\mathcal{S}$) to construct a subset of negative samples $\boldsymbol{y}^{-}_{t}=\{y_{t,k}^{-}\}_{k=1}^{K}$ for each token $y_t$.
Then, we pass the $\boldsymbol{y}^{-}$ through an embedding lookup table for negative embeddings $\boldsymbol{h}^{-}_{t}=\{h_{t,k}^{-}\}_{k=1}^{K}$.
\\
\textbf{Distance Metrics}. Previous works \cite{gao2021simcse,yan-etal-2021-consert,zhang2022frequency} use cosine function to calculate the similarity between positive and negative pairs.
Instead, we adopt KL-divergence as the distance metrics of contrastive learning because we believe that KL loss is more stringent compared to cosine loss (\textit{i.e.},
when the cosine loss of two vectors is 1, KL loss is not always 0).
For example, given two vectors $a=[1, 2, 3]$ and $b=[20, 40, 60]$, then $\operatorname{cos}(a,b)=1$, but $\operatorname{KL}(a,b)=4.04$.
Experimental results in Table \ref{tab:diff_cl} show the effectiveness of our method. 

The objective of contrastive learning is to minimize the following loss:
\begin{equation}
\mathcal{L}_{\mathrm{CL}}=-\sum_{y \in\boldsymbol{D}} \sum_{t \in T_y} \log \frac{e^{\phi\left(h_{t}, h_{t}^{+} \right) / \tau}}{\sum_{k \in K} e^{\phi \left(h_{t}, h^{-}_{t,k} \right) / \tau}},
\label{eq11}
\end{equation}
\begin{equation}
\phi\left(h_{t}, h_{t}^{+} \right) = \frac{1}{2}(\operatorname{KL}(u(h_{t}) \| u(h_{t}^{+}))+\operatorname{KL}(u(h_{t}^{+}) \| u(h_{t}))),
\label{eq12}
\end{equation}
where $\tau$ is a temperature hyperparameter and $u(\cdot)$ is $\operatorname{softmax}$ operation. 
The softmax function is applied to obtain the embedding distributions.
Actually, inspired by recent work \cite{li2021contrastive,zbontar2021barlow} that treats each dimension of an embedding as a latent class, above embedding distributions can therefore be considered as probability distributions of the latent classes.
Then, we calculate the bidirectional KL-divergence as the KL-divergence is a non-symmetric measure.

Finally, ConSLT can be optimized by jointly minimizing the traditional SLT loss and contrastive loss:
\begin{equation}
\mathcal{L}=\mathcal{L}_{\mathrm{SLT}}+\alpha\mathcal{L}_{\mathrm{CL}},
\label{eq13}
\end{equation}
where $\alpha$ is the weight to balance the two training losses.

\section{Experiments}

\subsection{Experimental Settings}
\textbf{Dataset and Metrics}. We evaluate our model on the German sign language dataset (PHOENIX14T) \cite{camgoz2018neural} and the Chinese sign language dataset (CSL-Daily) \cite{zhou2021improving}.
For fair comparisons, we follow \cite{camgoz2020sign} to construct a word-level vocabulary for texts.
We apply detokenized BLEU and Rouge-L F1 (ROUGE) to measure the translation quality \footnote{https://github.com/neccam/slt/blob/master/signjoey/metrics.py}.
\\
\textbf{Implementation Details}.
To verify the effectiveness of ConSLT, we examine it in two different settings.
For the Sign2Text setting, ConSLT consists of 3 encoder layers and 3 decoder layers, with each layer having 512 hidden units and 8 attention heads.
The batch size, initial learning rate, and dropout rate are set to 32, 1e-3, and 0.4, respectively.
For the Sign2Gloss2Text setting, both the encoder and decoder of STMC-Transformer have 2 layers, and the hidden units and attention heads are set to 512 and 8, respectively.
The batch size, initial learning rate, and dropout rate are set to 2048, 0.6, and 0.3, respectively.

To optimize our model, we employ Adam optimizer \cite{kingma2014adam} with $\beta_1= 0.9$, $\beta_2=0.998$, a learning rate schedule, early stopping and a shared weight matrix for the input and output word embeddings of the decoder.
For the $\mathcal{L}_{\mathrm{CL}}$, the temperature $\tau$, contrastive loss weight $\alpha$, and the number of negative samples $K$ are set to 0.1, 0.5, and 500, respectively. 
All hyper-parameters are tuned on the PHOENIX14T development set.
ConSLT is trained on 1 TITAN 24 GB for 4 hours.

\begin{table}[!t]
\centering
\footnotesize
\begin{tabular}{lcccc}
\toprule[0.8pt]
\multirow{2}{*}{Model} & \multicolumn{2}{c}{Dev}         & \multicolumn{2}{c}{Test}        \\ \cmidrule[0.3pt](lr){2-3} \cmidrule[0.3pt](lr){4-5}
                       & ROUGE          & BLEU-4          & ROUGE          & BLEU-4          \\ \cmidrule[0.5pt](lr){1-5}
\multicolumn{5}{l}{\textit{End-to-end: Sign2Text}}                                                     \\
RNN-SLT \cite{camgoz2018neural} & 31.80          & 9.94           & 31.80          & 9.58           \\
PiSLTRc \cite{xie2021pisltrc}   & \textbf{47.89}          & \textbf{21.48} & \textbf{48.13}          & 21.29          \\ \cmidrule[0.5pt](lr){1-5}
SL-Transformer\cite{camgoz2020sign} & 44.40          & 20.69          & 44.89          & 20.17          \\
ConSLT                 & 47.74 & 21.11          & 47.69 & \textbf{21.59}$^{\dagger}$ \\ \toprule[0.8pt]
\multicolumn{5}{l}{\textit{Cascading: Sign2Gloss2Text}}                                               \\
RNN-SLT \cite{camgoz2018neural}               & 44.14          & 18.40          & 43.80          & 18.13          \\
SL-Transformer \cite{camgoz2020sign}         & -              & 22.11          & -              & 22.45          \\ \cmidrule[0.5pt](lr){1-5}
STMC-Transformer \cite{yin-read-2020-better}       & 46.31          & 22.47          & 46.77          & 24.00          \\
ConSLT                 & \textbf{47.52} & \textbf{24.27} & \textbf{47.65} & \textbf{25.48}$^{\dagger}$ \\
\bottomrule[0.8pt]
\end{tabular}
\caption{Results on PHOENIX14T. $^{\dagger}$ denotes the improvement over our backbone networks is statistically significant ($p<0.05$).} 
\label{tab:main}
\end{table}
\begin{table}[!t]
\centering
\footnotesize
\begin{tabular}{lcccc}
\toprule[0.8pt]
\multirow{2}{*}{Model} & Dev            & \multicolumn{1}{l}{} & \multicolumn{2}{c}{Test}        \\ \cmidrule[0.3pt](lr){2-3} \cmidrule[0.3pt](lr){4-5}
                       & ROUGE          & BLEU-4                & ROUGE          & BLEU-4          \\ \cmidrule[0.5pt](lr){1-5}
SL-Transformer \cite{camgoz2020sign}         & 35.87          & 11.18                & 36.11          & 11.62          \\
ConSLT                 & \textbf{41.46} & \textbf{14.8}        & \textbf{40.98} & \textbf{14.53}$^{\ddagger}$ \\
\bottomrule
\end{tabular}
\caption{Results on CSL-Daily. $^{\ddagger}$ denotes the improvement over SL-Transformer is statistically significant ($p<0.01$).} 
\label{tab:main_csl}
\end{table}

\subsection{Main Results}
We compare ConSLT to previous methods on the two benchmarks PHOENIX14T and CSL-Daily, as shown in Table \ref{tab:main} and Tabel \ref{tab:main_csl}. 
The results are averages of ten runs with different random seeds.
ConSLT substantially improves translation quality for SLT in both the end-to-end and cascaded settings. 
In particular, ConSLT outperforms the baseline SL-Transformer  on PHOENIX14T and CSL-Daily by +1.42/+3.34 BLEU/ROUGE and +2.91/+5.59 BLEU/ROUGE, respectively, in the end-to-end setting.
In the cascaded scenario, ConSLT surpasses the baseline STMC-Transformer on PHOENIX14T by +1.48 BLEU and +1.21 ROUGE.
Experiments demonstrate that introducing contrastive learning can bring significant performance gains for SLT. 
Moreover, the cascaded SLT systems outperform the end-to-end methods by a large margin since the cascaded  systems use gloss as the intermediate supervised signals to avoid the long-term dependency issues \cite{camgoz2020sign}.

\begin{table}[!tbp]
\centering
\footnotesize
\begin{tabular}{l|cccc}
\toprule
              & BLEU-1    & BLEU-2    & BLEU-3    & BLEU-4    \\ \cmidrule(lr){1-5}
baseline (w/o CL)        & 48.73  & 36.53  & 29.03  & 24.00  \\
w/ S-CL $+$ cos & 48.48  & 37.17  & 29.49  & 24.36  \\
w/ T-CL $+$ cos & 49.91  & 37.60  & 29.88  & 24.59  \\ 
w/ S-CL $+$ KL  & 50.82  & 38.04  & 30.23  & 25.07  \\ 
w/ T-CL $+$ KL  & \textbf{51.57} & \textbf{38.81} & \textbf{30.91} & \textbf{25.48} \\ \bottomrule
\end{tabular}
\caption{Ablation study on the PHOENIX14T test set. }
\label{tab:diff_cl}
\end{table}

\subsection{Ablation Study}
We conduct the ablation study on the PHOENIX14T in the cascaded setting to investigate the influence of contrastive learning for SLT. 
First, following SimCSE \cite{gao2021simcse}, we train a model with a sentence-level contrastive learning framework. 
More specifically, we pass the sign gloss predicted by STMC network \cite{zhou2020spatial} to the vanilla  Transformer \cite{vaswani2017attention} twice and obtain two sentence representations as positive pairs. 
We treat other sentence representations within a mini-batch as negative examples. 
To make the comparisons fair, we apply the cosine function and KL-divergence to calculate the distance of positive and negative pairs for the ablation studies, respectively.
We set the number of negative samples to 500. 
For notation, \textbf{w/o CL} means no contrastive learning method, \textbf{S-CL} means sentence-level contrastive learning, \textbf{T-CL} means our token-level contrastive learning. 
As shown in Table \ref{tab:diff_cl}, we can see that the different contrastive learning methods used in SLT all bring significant improvements. 
Furthermore, ConSLT gains improvements of +0.41 BLEU, compared with sentence-level contrastive learning.
This indicates that ConSLT is more suitable for SLT as it can obtain fine-grained token representations.
Moreover, ConSLT which adopts KL-divergence as the distance metric of contrastive learning is better than that uses cosine function. 
As shown in the previous example (see Section \ref{sec:tcl}), we speculate that a possible reason is that learning similar representations via KL-divergence is more difficult than via cosine function.
The former requires the model to learn more distinguishable token representations (Figure \ref{fig:visual_conslt}). 
In summary, these ablation studies demonstrate the effectiveness of ConSLT.

\subsection{Negative Sampling Strategies}
We conduct comparative experiments in a cascaded setting to analyze the effectiveness of negative sampling strategies in contrastive learning.
Four sampling strategies were explored, as shown in Table \ref{tab:sampling}. 
For each token, \textbf{Batch} and \textbf{Vocab} randomly sample negative samples from a mini-batch and the vocabulary, respectively.
\textbf{Batch$\backslash$Sent} and \textbf{Vocab$\backslash$Sent} randomly take negative samples from a mini-batch and the vocabulary, respectively, but not in the current sentence.

The results show that \textbf{Vocab} outperforms \textbf{Batch} since the tokens in the vocabulary are more diverse than those in a mini-batch.
Furthermore, \textbf{Vocab$\backslash$Sent} performs better than other strategies. 
This result indicates that tokens in the same sentence should be semantically similar due to the contextual relationship, which is similar to predicting the masked tokens based on its contextual information in the masked language model \cite{devlin-etal-2019-bert}. 
It also enhances the relationship between the masked token and its context.
Therefore, different tokens from the same sentence cannot be pushed apart.
These comparative experiments also demonstrate the superiority of the proposed sampling strategy.

\begin{table}[!tbp]
\centering
\small
\begin{tabular}{l|cccc}
\toprule
           & BLEU-1    & BLEU-2    & BLEU-3    & BLEU-4            \\ \cmidrule(lr){1-5}
Batch      & 50.69  & 37.93  & 30.09  & 24.79          \\
Batch$\backslash$Sent & 50.84 & 38.22 & 30.24 & 24.92 \\
Vocab      & 51.20  & 38.33  & 30.45  & 25.14          \\
Vocab$\backslash$Sent & \textbf{51.57} & \textbf{38.81} & \textbf{30.91} & \textbf{25.48} \\ \bottomrule
\end{tabular}
\caption{BLEU scores with different sampling strategies of negative samples on the PHOENIX14T test set.}
\label{tab:sampling}
\end{table}

\section{Conclusion}
In this paper, we provide a new insight to alleviate the low-resource problem of SLT from the perspective of representation learning.
We introduce ConSLT, a token-level contrastive learning framework for SLT that aims to learn effective token representations by pushing apart tokens in the vocabulary that are outside the current sentence.
It is worth mentioning that ConSLT can be applied to different model structures.
We also explore the impact of various contrastive strategies and provide a fine-grained analysis to interpret how our approach works.
Experimental results demonstrate that contrastive learning can significantly improve translation quality for SLT.
In the future, we will further examine cross-modal relationships between sign videos and spoken language texts.

\section{Acknowledgements}
We thank all the anonymous reviewers for their insightful and valuable comments. This work was supported by National Natural Science Foundation of China "Research on Neural Chinese Sign Language Translation Methods Integrating Sign Language Linguistic Knowledge" (No. 62076211) and Central Leading Local Project "Fujian Mental Health Human-Computer Interaction Technology Research Center" (No. 2020L3024). 

\vfill\pagebreak
\begin{spacing}{0.9}
\bibliographystyle{IEEEbib}
\bibliography{refs}
\end{spacing}
\end{document}